%% file: main.tex
\documentclass[lettersize,journal]{IEEEtran}
\usepackage{amsmath,amsfonts}
\usepackage{algorithm}
\usepackage{algpseudocode}
\usepackage{array}
\newcommand{\thickhline}{\noalign{\hrule height 0.8pt}}
\usepackage[caption=false,font=normalsize,labelfont=sf,textfont=sf]{subfig}
\usepackage{float}
\usepackage{textcomp}
\usepackage{stfloats}
\usepackage{makecell}
\usepackage{multirow}
\usepackage{url}
\usepackage{amsthm}
\usepackage{tikz}
\usepackage{verbatim}
\usepackage{graphicx}
\usepackage{cite}
\usepackage{algpseudocode}
\usepackage[normalem]{ulem}

\begin{document}
\title{Meta-learning enhanced adaptive robot control strategy for automated PCB assembly}
\author{\IEEEauthorblockN{Jieyang Peng\IEEEauthorrefmark{1}\IEEEauthorrefmark{2}, Dongkun Wang\IEEEauthorrefmark{1}, Junkai Zhao\IEEEauthorrefmark{2}, \\Yunfei Teng\IEEEauthorrefmark{3}, Andreas Kimmig\IEEEauthorrefmark{2}, Xiaoming Tao\IEEEauthorrefmark{1}\IEEEauthorrefmark{4}, Jivka Ovtcharova\IEEEauthorrefmark{2}}

\IEEEauthorblockA{\IEEEauthorblockA{\IEEEauthorrefmark{2} Karlsruhe Institute of Technology, 76131 Karlsruhe, Germany}\\
\IEEEauthorblockA{\IEEEauthorrefmark{3}Engineer School, New York University, New York, NY 10012, USA}\\
\IEEEauthorrefmark{1}Department of Electronic Engineering, Tsinghua University, Beijing 100084, PR China}\\
\thanks{
\emph{This work was accepted as \uline{Journal of Manufacturing Systems 78 (2025) 46–57} without a proper review of the final version. After we notified the journal of these issues, the journal declined to correct the published record. With support from \uline{Tsinghua University’s Academic Committee}, this revised version accurately reflects the paper’s contributions and academic status.}\\
\IEEEauthorrefmark{3} Meta-method design and complete the whole methodology parts.
Corresponding to E-mail address: yt1208@nyu.edu (Y. Teng)\\
\IEEEauthorrefmark{4} Experiment supervision and advising for the remaining sections.
Corresponding to E-mail address: taoxm@tsinghua.edu.cn (X. Tao).}}

\maketitle
\input{abstract}
\input{introduction}

\input{related}

\input{methodology}
\input{system}

\input{experiment}
\input{conclusion}

\bibliographystyle{IEEEtran}
\bibliography{ref}

\end{document}

%% file: abstract.tex
\begin{abstract}
The assembly of printed circuit boards (PCBs) is one of the standard processes in chip production, directly contributing to the quality and performance of the chips. In the automated PCB assembly process, machine vision and coordinate localization methods are commonly employed to guide the positioning of assembly units. However, occlusion or poor lighting conditions can affect the effectiveness of machine vision-based methods. Additionally, the assembly of odd-form components requires highly specialized fixtures for assembly unit positioning, leading to high costs and low flexibility, especially for multi-variety and small-batch production. Drawing on these considerations, a vision-free, model-agnostic meta-method for compensating robotic position errors is proposed, which maximizes the probability of accurate robotic positioning through interactive feedback, thereby reducing the dependency on visual feedback and mitigating the impact of occlusions or lighting variations. The proposed method endows the robot with the capability to learn and adapt to various position errors, inspired by the human instinct for grasping under uncertainties. Furthermore, it is a self-adaptive method that can accelerate the robotic positioning process as more examples are incorporated and learned. Empirical studies show that the proposed method can handle a variety of odd-form components without relying on specialized fixtures, while achieving similar assembly efficiency to highly dedicated automation equipment. As of the writing of this paper, the proposed meta-method has already been implemented in a robotic-based assembly line for odd-form electronic components. Since PCB assembly involves various electronic components with different sizes, shapes, and functions, subsequent studies can focus on assembly sequence and assembly route optimization to further enhance assembly efficiency.

\end{abstract}
\vspace{-0.2in}

%% file: introduction.tex
\section{Introduction}
In the context of globalization, two prevailing opinions have emerged in the manufacturing industry. Firstly, decreasing production costs and production cycle times is essential when faced with peer competition \cite{1}. Secondly, due to the increasing demand for individualized production, manufacturing enterprises should have the rapid response capability for personalized demand. Consequently, companies nowadays need to develop assembly systems that are flexible enough to assemble new products with minimal reconfiguration \cite{2}. This is most evident in the electronics industry, where the dominant activity has become Printed Circuit Board (PCB) assembly. As a result, electronic manufacturers are faced with the challenge of producing individualized products economically. Against this background, flexibility and adaptability are becoming decisive success factors for production systems \cite{3}.

As the dominant procedure of the PCB production line, PCB assembly is usually accomplished by robots with machine vision or coordinate positioning systems. Robotic positioning errors can be broadly categorized as internal errors and external errors \cite{4}. Internal errors arise from factors such as joint and link compliance or gear backlash associated with the robot. On the other hand, external errors result from factors in the robot's external environment, such as inaccurate tool and fixture positions.

To enhance the accuracy of robotic positioning, auxiliary devices, such as positioners, have been developed to guide the robotic end effector to the desired point with increased precision \cite{5}. Nonetheless, these auxiliary devices are typically designed for a specific task and are not easily adaptable to other tasks. This can lead to high reconfiguration costs and low production system flexibility, which is extremely detrimental when facing individualization challenges.

Vision-based robotic positioning mainly involves capturing visual data through a camera, which can provide either 2D or 3D information, and subsequently leveraging this information to infer the object's geographical location \cite{6}. Although numerous studies have focused on enhancing the position accuracy of vision-based position control \cite{7,8}, this approach inherently suffers from limitations, e.g., unexpected occlusions or poor lighting conditions. Given the challenges in accurately depicting positioning errors, artificial neural networks offer a viable solution for compensating the absolute positioning errors \cite{9,10}. The performance of neural networks is highly dependent on the quality of training data, which is difficult to acquire from real-world robotic environments. The structure and parameters of neural networks significantly impact their performance. Therefore, it is a challenging task to construct a neural network that can effectively generalize to various position errors encountered in real-world scenarios. In complicated robotic applications (e.g., robotic assembly), the trained neural network may not effectively generalize to new tasks without fine-tuning.

In complicated robotic applications, such as robotic assembly in a cluttered environment, occlusions between manipulated objects can influence the end-effector's positioning ability \cite{11}. Considering the vision's inherent limitations in occlusions, a vision-free positioning method is proposed. The robotic positioning problem is first comprehensively addressed from a statistical standpoint by formulating it as a probability maximization task via sampling from a Gaussian distribution, which is consistent with the principles of the law of large numbers, resulting in a complete and effective solution. A numerical method adapted from the Monte Carlo method is proposed to improve sampling accuracy, which serves to approximate the integral. To evaluate the effectiveness of the proposed method, printed circuit board (PCB) assembly was chosen as the study case, where the robot assembles a set of non-standard odd-form electronic components onto the PCBs.

The proposed method addresses the visual challenges posed by inadequate lighting or occlusion by leveraging a model-agnostic metamethod that compensates for robotic position errors through interactive feedback. Unlike traditional vision-based positioning, which relies heavily on clear and consistent visual data, this method endows the robot with the capability to learn and adapt to various positioning errors. By formulating the positioning problem as a probability maximization task and utilizing statistical principles, the method can accurately infer the object's geographical location even in conditions where visual data is limited or obscured. Furthermore, the self-adaptive nature of the meta-method allows the robot to continually improve its positioning accuracy as more examples are incorporated and learned, making it highly resilient to the inherent limitations of vision-based systems. Thus, the proposed method offers a viable solution to the challenges posed by inadequate lighting or occlusion in robotic positioning tasks.

The main contributions of this work are listed as follows:

\noindent\textbf{(1) Design of a novel self-adaptive meta search method for robotic positioning.} The meta-method involves two key steps:

\noindent\textbf{\textit{(a)}} \underline{Updating the fitted posterior of the Gaussian distribution}

\noindent\textbf{\textit{(b)}} \underline{Maximizing the rate of successful sampling based on the} \underline{modeled distribution.} The probability of accurate positioning will be maximized through interactive feedback.

\noindent\textbf{(2) Design of an automated assembly facility for odd-form electronic components.} The automatic assembly equipment for odd-form electronic components was designed to enable the assembly of various types of odd-form electronic components. Collaborative robots are utilized for the assembly. A special gripper has been designed to ensure the smooth clamping of odd-form electronic components.

\noindent\textbf{(3) Empirical study of the proposed method.} An empirical study of the assembly efficiency is conducted with a typical odd-form component as the experimental object. The experimental results indicate that the proposed meta search method achieves superior assembly efficiency compared with other methods, while also possessing the best self-adaption ability.

%% file: related.tex
\section{Related work}
The current PCB assembly positioning methods primarily involve machine vision and coordinate positioning. This chapter briefly describes the applications of these two methods, analyzes their usage scenarios and limitations, and finally summarizes the difficulties of PCB assembly positioning.

\subsection{PCB assembly based on machine vision}
Machine vision-based methods have been extensively employed in PCB assembly processes, particularly in scenarios with stable light sources, due to their adaptability and flexibility in overcoming tolerances related to chip placement, shape, and assembly unit positioning \cite{12}. These methods leverage advanced image processing techniques to provide detailed positional information, guiding robotic systems with high precision.

Fontana \cite{13} presented a pioneering experimental setup utilizing a 4-degree-of-freedom robot and a dual-camera vision system. This setup demonstrated the effectiveness of vision-based strategies in supporting precise manipulation operations, with a portable and flexible program integrating both machine vision and control modules. Similarly, Lin \cite{14} developed a machine vision-based system for inserting odd-form electronic components, leveraging dual SCARA robots for enhanced flexibility and productivity. Zhang \cite{15} proposed a method combining threshold segmentation, feature extraction, and edge contour fitting to detect and locate odd-form components and PCB holes, addressing the challenge of edge unsmoothness through contour fitting. Liang \cite{16} introduced a rotational stereo vision system for autoinsertion, enabling precise pose detection of electronic components. Liu et al. \cite{17} presented robust image enhancement, edge detection, and feature extraction techniques for component detection and localization, employing histogram key point constrained homomorphic filtering and line/arc segment detection. Nerakae \cite{18} integrated a machine vision system with a robotic system for pick-and-place operations, using image processing software to control a SCARA robot based on the shape and orientation of the assembling space.

Despite their widespread use, machine vision-based methods face several limitations. Camera calibration deviation can compromise the reliability of location output, leading to potential errors in assembly processes \cite{19}. Additionally, visual occlusion and confusion pose significant challenges, particularly when dealing with large or intricately shaped electronic components. The deployment of complex vision systems can also be cost-prohibitive, limiting their feasibility in multi-variety and small-batch production scenarios.

In contrast to these vision-based methods, the proposed vision-free, model-agnostic meta-method for compensating robotic position errors offers several distinct advantages. By maximizing the probability of accurate robotic positioning through interactive feedback, this method reduces the dependency on visual feedback and mitigates the impact of occlusions or lighting variations. Furthermore, it endows the robot with the capability to learn and adapt to various position errors, inspired by human instincts for grasping under uncertainties. This selfadaptive nature accelerates the robotic positioning process as more examples are incorporated and learned, leading to improved efficiency and flexibility.

Empirical studies have shown that the proposed method can handle a variety of odd-form components without relying on specialized fixtures, achieving similar assembly efficiency to highly dedicated automation equipment. This advantage is particularly significant in multivariety and small-batch production environments, where the cost and flexibility of vision-based systems become limiting factors.

\subsection{PCB assembly based on coordinate positioning}
The assembly of printed circuit boards (PCBs) is a vital process in electronics manufacturing, with coordinate positioning being a commonly employed method, particularly in standardized production lines. This approach involves the precise placement of electronic components onto a PCB using predefined coordinates, facilitated by advanced machinery and software. Sassanelli et al. \cite{20} highlight the reliance on accurate machinery and software programming to control the placement heads, ensuring each component is aligned and fixed at its designated position. This method leverages digital models and simulations for process design and programming, offering flexibility and the capacity to address uncertainties during execution.

However, despite its widespread use, the coordinate positioning method has limitations, particularly when dealing with odd-form components. Mathiesen \cite{21} presents a simulation-based assembly platform specifically for through-hole technology (THT) components, demonstrating the use of digital models and simulations to increase flexibility. In this context, Mauro Queirós \cite{22} proposes and validates a collaborative robotic work cell with vision systems to autonomously or semi-autonomously insert pin through-hole components (PTH) in PCB assembly, replacing the need for manual insertion. Nevertheless, the need for highly specialized equipment remains a significant challenge, especially when considering the diverse range of component shapes and sizes encountered in PCB assembly. Xu \cite{23} introduces a compliant wrist design for robot manipulators, which provides flexibility to correct positioning errors and avoid high impact forces. While this addresses some of the limitations of rigid positioning systems, it does not fully mitigate the need for specialized fixtures.

Furthermore, achieving high precision in coordinate positioning is crucial, as variation in pallet position can significantly impact assembly quality and yields. Vallance \cite{24} proposes an exact constraint approach using a split-groove kinematic coupling to reduce variation in pallet location, improving precision. Similarly, Liu \cite{25} presents a highaccuracy pose measurement system designed to enhance the accuracy of automated robotic assembly, utilizing two EPSON C4-A601S robots, vision and force controllers, and PLCs to achieve high-precision. Additionally, Xurui Li \cite{26} explores a different approach by presenting a Digital Twin (DT) model for automating PCB assembly, using a symmetry-driven method and a three-stage learning approach to optimize robotic assembly. These advancements underscore the ongoing efforts to improve precision in coordinate positioning, yet they still rely heavily on dedicated equipment and precise hardware accuracy.

The proposed vision-free, model-agnostic meta-method for compensating robotic position errors offers a significant advantage. By maximizing the probability of accurate robotic positioning through interactive feedback, this method reduces dependency on visual feedback and mitigates the impact of occlusions or lighting variations. Its selfadaptive nature allows the robot to learn and adapt to various position errors, inspired by human instincts for grasping under uncertainties. This capability enables the method to handle a variety of odd-form components without relying on specialized fixtures, thereby enhancing production line flexibility and reducing costs.

Moreover, the proposed method's ability to accelerate the robotic positioning process as more examples are incorporated and learned represents a significant advancement over traditional coordinate positioning methods. Unlike existing approaches that require extensive calibration and specialized equipment, the proposed meta-method leverages interactive feedback and learning to continuously improve positioning accuracy. This makes it particularly suitable for multi-variety and small-batch production, where the flexibility and adaptability of the production line are paramount.

%% file: methodology.tex
\section{Methodology}
\subsection{Problem statement and overall methodology}
In the electronic component assembly process, the printed circuit board (PCB) is initially mounted onto a tray and securely fastened \cite{27}, after which the end effector carries out a series of operations. First, the end effector picks up the electronic component using the fixture. Next, the end effector moves the component to a predetermined insertion position. Finally, the end effector inserts the component into the circuit board, completing the assembly process. Accurate positioning of the electronic component during insertion is of utmost importance, as any inaccuracies can lead to failure. The robot may encounter dynamic constraints or occlusions caused by multiple electronic components previously inserted into the PCB. Moreover, the time required for the insertion process is also crucial, as an excessively long insertion time can render the process impractical. Hence, it is crucial for the robot to perform the assembly process with maximum precision and efficiency.

\begin{figure}[!t]
\centering
\includegraphics[width=1\linewidth]{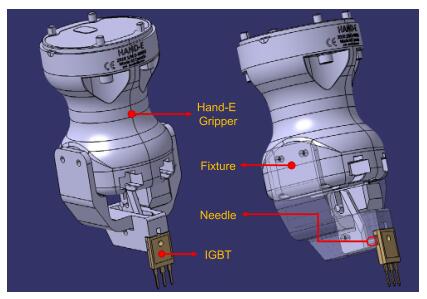}
\caption{Illustration of robot gripper (in virtual environment).}
\label{f1}
\end{figure}

One typical odd-form electronic component, the IGBT (Insulated Gate Bipolar Transistor) \cite{28}, is selected for assembly in the PCB. The IGBT's appearance is displayed in Fig. \ref{f1}. This type of component is challenging for automatic assembly; each IGBT possesses three pins and can only be successfully inserted when all three pins are simultaneously inserted into their respective holes. To pick up the IGBT, a 3D-printed fixture mounted on the gripper is developed. The opening and clamping actions of the fixture are achieved by the parallel movement of two sliders on the gripper. During the fixture operation, the sliders move in opposite directions. The pickup mechanism within the fixture completes the component's pickup action. During the component pickup process, one slider moves in a certain direction, and the fixture controls the slider's movement. Through the pickup mechanism on the gripper, the metal needle rotates along the clamping direction of the IGBT. When the metal needle rotates into the through-hole in the IGBT and clamps it, the slider of the fixture stops moving. When the IGBT is successfully inserted into the circuit board, the slider of the fixture moves in the opposite direction, and the metal needle also rotates in the opposite direction. The metal needle rotates out of the through-hole and releases the IGBT.

\begin{figure}[!t]
  \includegraphics[width=0.5\textwidth]{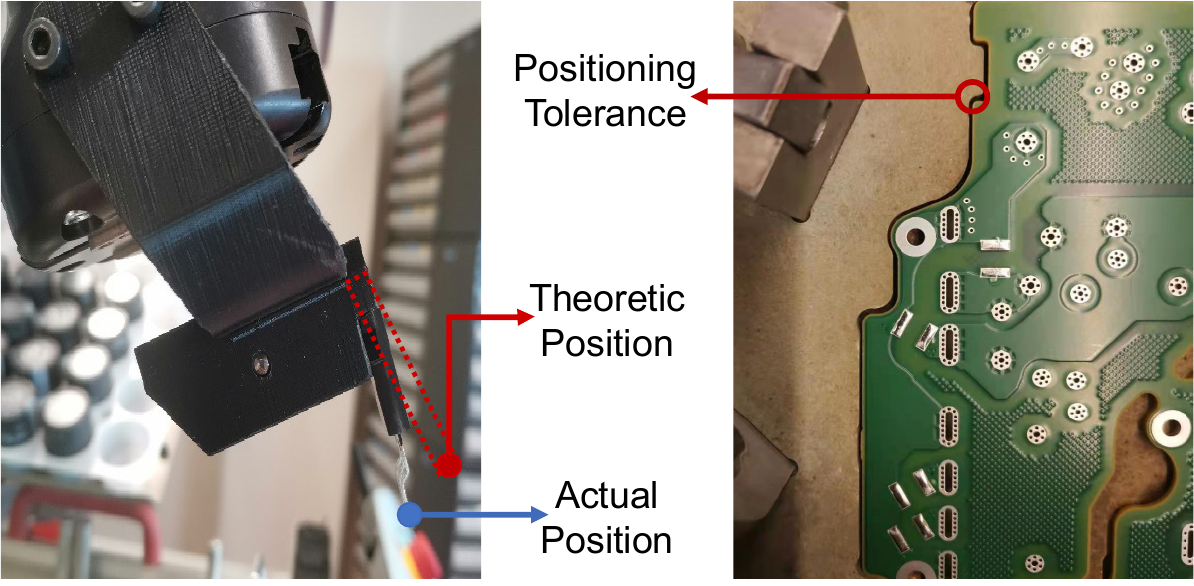}
  \caption{Examples of components clamping tolerance(left) and positioning tolerance(right) during assembly.}
  \label{f2}
\end{figure}

The primary factor that impacts the precision of the assembly is the range of tolerances involved. Various tolerances affect the robot's positioning ability in the assembly process. For example, shape tolerance arises from the bending of the pin due to its low stiffness, while component clamping tolerance arises when the fixture fails to clamp the component accurately. Additionally, the PCB's manufacturing tolerance and the tray's position tolerance can also affect the hole's position on the PCB. An example of components clamping tolerance and PCB position tolerance is shown in Fig. \ref{f2}.

\begin{figure*}[!tbh]
  \centering
  \includegraphics[width=0.9\linewidth]{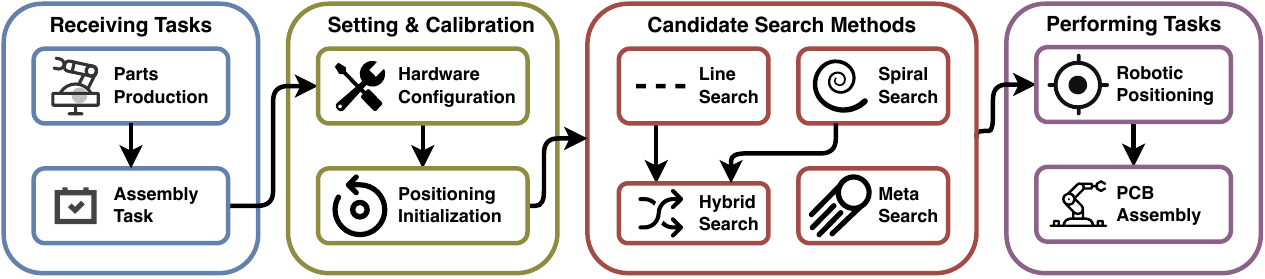}
  \caption{The overview experiment process.}
  \label{f3}
\end{figure*}

A tolerance-adaptive assembly method will be developed to handle the various tolerances listed above. This paper focuses on the adaptive positioning problem during the assembly process and the effectiveness of the proposed mathematical model in addressing it. The basic principles involved are as follows. Upon grasping an IGBT, the robot moves it to the predetermined insertion position and proceeds to perform the insertion operation. The initial insertion of the IGBT is likely to fail due to various tolerances. When the pins of the IGBT cannot be inserted into the circuit board, there is a reaction force between the electronic pins and the circuit board, according to which the robot determines whether the components are successfully assembled through the integrated force sensor. In case of an assembly failure, the robotic system will responsively initiate a search process near the intended insertion position. Analogous to the fine-tuning in the manual assembly process, the robot will explore the actual insertion positions until the IGBT's pins are inserted into the mounting holes. The coordinates of the successful assembly position will be recorded in the system, allowing the model to learn from these successful attempts and adapt to future tasks. The overall method of this paper is shown in Fig. \ref{f3}.

To verify the effectiveness of the proposed mathematical model, an empirical study was conducted with a typical odd-form component as the experimental object. The experimental setup included a robotic arm equipped with the proposed model-agnostic meta-method for positioning error compensation. The model was trained using a dataset of successful assembly positions and was then tested on a series of insertion tasks. The results showed that the proposed model significantly improved the accuracy of robotic positioning compared to baseline methods that did not use the model. Additionally, the model demonstrated its ability to adapt to new assembly tasks and different component tolerances.

\subsection{Mathematical modeling}
The emphasis is on addressing position errors in robots employed for assembly tasks using a mathematical model that captures the uncertainty in positioning. This study commences with establishing such a model and providing a rigorous explanation of the algorithms that underpin it. The model is based on a Gaussian distribution assumption, which allows for the representation of uncertainty in positioning and facilitates the iterative updating of the posterior distribution based on past observations.

To further validate the effectiveness of the proposed mathematical model, a comparison was made with other models and methods for robotic positioning. The results showed that the proposed model outperformed these alternative methods in terms of accuracy and adaptability to new assembly tasks. This superiority was attributed to the model's ability to learn from past experience and adapt to different 
component tolerances, as well as its reliance on interactive feedback for continuous improvement.

\begin{figure}[!t]
\centering
\includegraphics[width=1\linewidth]{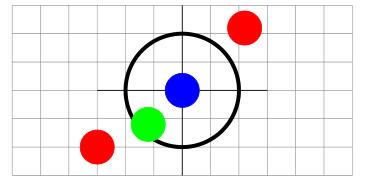}
\caption{Cases of component insertion. {\bf Blue circle}: feasible; {\bf Red circle}: not feasible;
{\bf Green circle}: on the boundary. (For interpretation of the references to color in this
figure legend, the reader is referred to the web version of this article.)}
\label{f4}
\end{figure}

The robot assembly process involves retrieving an object by the robot and subsequently inserting it into another object at a predetermined position. As depicted in Fig. \ref{f4}, the enclosed largest black circle represents the feasible area around the insertion point (center of the largest circle). The object can be successfully assembled when inserted into the feasible area.

It should be noted that each attempt to assemble the components may result in a slightly different feasible insertion area. Thus, the center of the feasible area is considered a random variable $\mathbf{a}$ and the radius of that area is considered a random variable $\mathbf{b}$, which are assumed to be random variables with bounded mean and variance (i.e., $\|\mathbf{a}\|+\|\mathbf{b}\|< +\infty$ ). In short, the distribution of the feasible area can be represented as:
\begin{equation}
\mathbf{c}=\mathbf{a}+\mathbf{b} \quad \mathbf{a} \sim \mathcal{N}\left(\mu_{a}, \Sigma_{a}\right), \mathbf{b} \sim \mathcal{N}\left(\mu_{b}, \Sigma_{b}\right) 
\end{equation}
where $\mathbf{c}$ is also a random variable. For simplicity, both $\mathbf{a}$ and $\mathbf{b}$ are assumed to be Gaussian random variables. Consequently, $\mathbf{c}$ is also a Gaussian random variable. In our following analysis, only one Gaussian distributed random variable is considered.

{\bf Proof.} The characteristic function of a multidimensional random variable is defined as the Fourier transform of its joint probability density function (PDF). Let us denote the multidimensional random variable as $\mathbf{X}=\left(X_{1}, X_{2}, \ldots, X_{n}\right)$, where $n$ is the number of dimensions. The characteristic function of $\mathbf{X}$, denoted as $\phi(\mathbf{t})$, is defined as:
\begin{equation}
\phi(\mathbf{t})=E\left[e^{i \mathbf{t} \cdot \mathbf{X}}\right] 
\end{equation}
where $\mathbf{t}=\left(t_{1}, t_{2}, \ldots, t_{n}\right)$ is a vector of parameters, $\mathbf{t} \cdot \mathbf{X}$ represents the dot product of the vectors $\mathbf{t}$ and $\mathbf{X}$, and $i$ is the imaginary unit.

The characteristic function of the sum of two independent random variables $X$ and $Y$ is simply the product of the two separate characteristic functions:
\begin{equation}
\varphi_{X+Y}(t)=E\left[e^{i \mathbf{t} \cdot(\mathbf{X}+\mathbf{Y})}\right]=\varphi_{X}(t) \cdot \varphi_{Y}(t)
\end{equation}
where:
\begin{equation}
\varphi_{X}(t)=E\left[e^{i \mathbf{t} \cdot \mathbf{X}}\right], \quad \varphi_{Y}(t)=E\left[e^{i \mathbf{t} \cdot \mathbf{Y}}\right]
\end{equation}

The characteristic function of the normal distribution with expected value $\mu$ and covariance matrix $\Sigma$ is:
\begin{equation}
\varphi(t)=\exp \left(i t^{T} \mu-\frac{t^{T} \Sigma t}{2}\right)  
\end{equation}

Thus:
\begin{align}
\varphi_{X+Y}(t) & =\varphi_{X}(t) \cdot \varphi_{Y}(t)\\
& \!=\!\exp \left(\!i t^{T} \mu_{X}\!\!-\!\!\frac{t^{T} \Sigma_{X} t}{2}\!\right)\!\! \exp \left(\!i t^{T} \mu_{Y}\!\!-\!\!\frac{t^{T} \Sigma_{Y} t}{2}\!\right) \\
& =\exp \left(i t^{T}\left(\mu_{X}+\mu_{Y}\right)-\frac{t^{T}\left(\Sigma_{X}+\Sigma_{Y}\right) t}{2}\right)
\end{align}

This is the characteristic function of the normal distribution with expected value $\mu_{X}+\mu_{Y}$ and covariance matrix $\Sigma_{X}+\Sigma_{Y}$.

Finally, recall that no two distinct distributions can both have the same characteristic function, so the distribution of $X+Y$ must be this normal distribution. 

The goal is to insert each object into the appropriate area successfully with minimal attempts. The probability of each attempt falling into the feasible area is being maximized, which in turn leads to the maximization of the integral problem. To be more specific, assuming that $n$ attempts can be performed, with the $i$ th attempt having tolerance $\epsilon_{i}$ centered at $x_{i}$, let $D_{i}$ be the circle of radius $\epsilon_{i}$ centered at $x_{i}$. The following equation will be resolved:
\begin{equation}
\max \int_{\cup_{i=1}^{n} D_{i}} f_{c}(y) d y 
\end{equation}
where $f_{c}(y)$ is the probability distribution of $\mathbf{c}$ at point $y$. The union operator is used instead of summation to avoid repeated sampling in the overlapping areas.

\subsection{Algorithmic-based search method}
This section commences by presenting several search methods characterized by a predefined sampling trajectory and a termination criterion based on the successful execution of the insertion task.

\subsubsection{Linear search}
The basic principle of the linear search method is that the robot will implement a linear search process along a straight search line to determine the hole position. The linear search method originates from the work by Chhatpar \cite{29} and has been modified to better suit PCB assembly requirements. The linear search method involves linear motion along a defined search line starting from the robot's predefined insertion position. The end-effector gripping a component will move to the predefined insertion position during the insertion process and perform the insertion step. If the robot successfully inserts the component into the hole, the robot will not continue the linear search method. If the robot cannot insert the peg, the end-effector will move along a search line at a specific distance to reach a new insertion position. The robot will then execute the insertion action from that new insertion position. Since the actual hole could be located randomly along the search line, the predefined insertion position is set as the central point of the search line. Theoretically, the actual insertion position could be uniformly distributed on either side of the central point. Therefore, the search positions on either side of the central point follow a uniform distribution. The end-effector moves in reverse search directions within two consecutive search cycles, which ensures that both directions from the predefined insertion position can be searched. The linear search process will be terminated upon successfully inserting the component into the hole or upon reaching the maximum allotted search cycles.

During the search process, the end-effector gripping the component moves along a certain search line $l$. In the $i$ th insertion cycle, the insertion position $P_{i}\left(x_{i}, y_{i}, z\right)$ follows:
\begin{equation}
x_{i}=x_{i-1}+s \cdot i \cdot(-1)^{i} \cdot \cos \theta 
\end{equation}
\begin{equation}
y_{i}=y_{i-1}+s \cdot i \cdot(-1)^{i} \cdot \sin \theta 
\end{equation}
where $\theta$ denotes the angle between the search line $l$ and the $X$ axis of the robot's base coordinate. The present study focuses on the assembly process in a two-dimensional plane, where the $z$ element of the insertion position remains constant for all insertion positions. $P_{0}\left(x_{0}, y_{0}, z\right)$ is the predefined insertion position. The insertion process is simplified by omitting the rotation movement of the peg. The distance between two successive insertion points is the stride $s$, and the maximum search cycles is $m$.

\subsubsection{Spiral search}
The end-effector's path during the spiral search process \cite{30} is an Archimedean spiral. The spiral search method is employed if the robot cannot insert the component at the predefined insertion position. In the spiral search process, the end-effector executes an Archimedean spiral motion originating from the predefined insertion position. This motion continuously expands outward on the $X Y$ plane. During each search step, the maximum search radius of the Archimedean spiral is $r$. Given $P_{0}$, the angular change between two successive insertion cycles is $\theta_{d}$, and the cumulative angular search range $\theta_{m x}$ is defined as:
\begin{equation}
\theta_{m x}=t \cdot 360^{\circ} 
\end{equation}
where $t$ is the total number of turns of the spiral. The per-degree change in radius $r_{d}$ is:
\begin{equation}
r_{d}=\frac{r}{\theta_{m x}} 
\end{equation}

In the $i$ th insertion cycle, the accumulated degrees the peg has obtained from the start of the search implementation, $\theta_{{cur }}$, is:
\begin{equation}
\theta_{c u r}=\theta_{d} \cdot i 
\end{equation}

The current radius $r_{{cur }}$ of the $i$ th search point on the spiral is calculated as:
\begin{equation}
r_{c u r}=\theta_{c u r} \cdot r_{d} 
\end{equation}

The $i$ th insertion position $P_{i}\left(x_{i}, y_{i}, z\right)$ is updated using the following equations:
\begin{equation}
x_{i}=x_{0}+r_{c u r} \cdot \cos \theta_{c u r} 
\end{equation}
\begin{equation}
y_{i}=y_{0}+r_{c u r} \cdot \sin \theta_{c u r} 
\end{equation}

During the spiral search process, the robot, with its compliant feature, consistently exerts a slight force in the $-Z$ direction to the component. This exerted force enables the component to be inserted into the hole as soon as it moves over the hole. Similarly to the linear search method, the spiral search process is terminated upon successfully inserting the component into the hole or reaching the maximum search radius defined by $r$.

\subsubsection{Hybrid search}
The drawback of the linear search method is that it will only succeed in locating the target hole if it is on the search line, while a hole deviating from the search line will not be reached. This requires the robot to conduct tedious search cycles step by step, which is timeconsuming. Compared with the linear search method, the spiral search method covers a more extensive area, making it more reliable for finding dispersively distributed holes. However, the search area is constrained within a limited region, and the robot using the spiral search method may fail to insert the component if the hole is far from the predefined insertion position.

To address these limitations, a hybrid search method that balances the drawbacks of these two search strategies is developed. The hybrid search method combines linear and spiral search strategies to improve the efficiency and reliability of the robotic positioning process.

The hybrid search method initiates with a linear search phase, adhering to a predefined trajectory in an attempt to pinpoint the insertion hole. This method is mathematically represented as a series of movements along a defined search line, starting from the robot's predefined insertion position. The position $P_{i}$ of the end effector during the $i$ th attempt is given by:
\begin{equation}
P_{i}=P_{i-1}+s \cdot(-1)^{i} \cdot \hat{d}
\end{equation}
where $s$ denotes the stride length, and $\hat{d}$ is the unit direction vector of the search line.

Should the linear search prove unsuccessful due to various tolerances, such as positional inaccuracies or component variances, the method seamlessly transitions to a spiral search. This phase commences from the predefined insertion point and expands radially outward on the XY plane, as defined by:
\begin{align}
r_{i} & =r_{i-1}+\Delta r  \\
\theta_{i} & =\theta_{i-1}+\Delta \theta  \\
P_{i, x} & =P_{i-1, x}+r_{i} \cdot \cos \left(\theta_{i}\right) \\
P_{i, y} & =P_{i-1, y}+r_{i} \cdot \sin \left(\theta_{i}\right)
\end{align}\\
where $r_{i}$ and $\theta_{i}$ represent the radius and angle of the spiral at the $i$ th step, respectively, and $\Delta r$ and $\Delta \theta$ are the incremental changes in radius and angle, respectively.

Algorithm 1 illustrates the complete hybrid search method. The algorithm begins by setting the search mode to linear search. The robot then performs a linear search along a predefined direction to locate the insertion hole. If the component is successfully inserted, the search process is terminated. If the component cannot be inserted, the search mode is switched to spiral search, and the robot performs a spiral search motion to locate the target hole. The algorithm continues to iterate until all components are successfully inserted.

\begin{algorithm}[ht]
\caption{Hybrid search method}
\begin{algorithmic}[1] 
\Require predefined insertion position
\While{component not inserted}
    \State search mode $\leftarrow$ linear search
    \If{component is inserted then}
        \State search mode $\leftarrow$ search break
    \ElsIf{component is not inserted then}
        \State search mode $\leftarrow$ spiral search
    \EndIf
\EndWhile
\Return actual insertion position
\end{algorithmic}
\end{algorithm}

\subsection{Meta search}
The aforementioned search methods cannot be adjusted flexibly, and these methods do not leverage useful information from previous samples at all \cite{31}. As a result, a new meta search method that iteratively updates the status of the learning model to speed up insertion processes is proposed.

The model is, in principle, divided into two stages: (1) a prior distribution is initially defined, and the posterior will be updated recursively, and (2) a search method based on gradient ascent is introduced. Based on the Gaussian distribution assumption, this update can be done recursively, meaning these two steps will be performed alternately.

\subsubsection{Gaussian distribution}
The formula for a two-dimensional Gaussian function \cite{32} in matrix form is as follows \cite{33}:
$$f(\mathbf{x})=\frac{1}{\sqrt{2 \pi|\Sigma|}} \exp \left(-\frac{1}{2}(\mathbf{x}-\boldsymbol{\mu})^{T} \Sigma^{-1}(\mathbf{x}-\boldsymbol{\mu})\right)$$

In this formula, $\mathbf{x}$ represents the vector of variables $(x, y), \boldsymbol{\mu}$ denotes the mean vector ($\mu_{1}, \mu_{2}$), and $\boldsymbol{\Sigma}$ represents the covariance matrix:
$$\boldsymbol{\Sigma}=\left[\begin{array}{cc}\sigma_{1}^{2} & \rho \sigma_{1} \sigma_{2} \\ \rho \sigma_{1} \sigma_{2} & \sigma_{2}^{2}\end{array}\right]$$

Here, $\sigma_{1}$ and $\sigma_{2}$ represent the standard deviations in the $x$ and $y$ directions, respectively, and $\rho$ denotes the correlation coefficient between the $x$ and $y$ variables.

\subsubsection{Distribution update}
The problem can be solved using Maximum Likelihood Estimation (MLE) \cite{34} or Maximum a Posteriori (MAP) \cite{35}. Firstly, it is noteworthy that a likelihood function can be expressed as:
$$\hat{\theta}_{M L E}(x)=\arg \max _{\theta} f(x \mid \theta)$$
while a posterior can be given by:
$$\hat{\theta}_{{MAP }}(x)=\arg \max _{\theta} f(x \mid \theta)=\arg \max _{\theta} f(x \mid \theta) g(\theta)$$

For Gaussian distribution, there is a closed-form solution for estimation. Taking the logarithm of the Gaussian distribution, the results are obtained as follows:
\begin{equation}
\log f(\mu, \Sigma ; x)  =-\frac{1}{2} \log |\Sigma|-\frac{1}{2}(x-\mu)^{T} \Sigma^{-1}(x-\mu)     
\end{equation}
\begin{equation}
\frac{\partial}{\partial \mu} \log f(\mu, \Sigma ; x)  =-\Sigma^{-1}(x-\mu)     
\end{equation}
\begin{equation}
\frac{\partial}{\partial \Sigma^{-1}} \log f(\mu, \Sigma ; x)  =-\frac{1}{2} \Sigma-\frac{1}{2}(x-\mu)(x-\mu)^{T}     
\end{equation}

By setting the gradients to zero, the results are:
\begin{equation}
 \hat{\mu}=\frac{1}{m} \sum_{i=1}^{m} x_{i},   
\end{equation}
\begin{equation}
\hat{\Sigma}=\frac{1}{m} \sum_{i=1}^{m}\left(x_{i}-\hat{\mu}\right)\left(x_{i}-\hat{\mu}\right)^{T} 
\end{equation}

Note that MLE is a biased estimator for the covariance matrix, but in practice, it does not make much difference as the number of data points increases. To solve for MAP, assuming that each element of both $\mu$ and $\Sigma$ is given a prior of Gaussian distribution $\mathcal{N}(0,1)$, then Eqs. (23), (24), and (25) become:
\begin{align}
\!\log\! f(\mu, \Sigma ; x)\!=\! & \!-\frac{1}{2}\! \log |\Sigma|\!\!-\!\!\frac{1}{2}(x\!\!-\!\!\mu)^{T} \Sigma^{-1}(x\!-\!\mu) \! \\
& \!-\!\frac{\lambda_{1}}{2} \cdot\|\mu\!-\!\tilde{\mu}\|_{2}^{2}-\frac{\lambda_{2}}{4} \cdot\|\Sigma\!-\!I\|_{2}^{2}\nonumber \\
\frac{\partial}{\partial \mu} \log f(\mu, \Sigma ; x)= & -\Sigma^{-1}(x-\mu)-\lambda_{1}(\mu-\tilde{\mu}) \\
\frac{\partial}{\partial \Sigma^{-1}} \log f(\mu, \Sigma ; x)= & \!-\!\frac{1}{2} \Sigma\!-\!\frac{1}{2}(x\!-\!\mu)(x\!-\!\mu)^{T}\!-\!\frac{\lambda_{2}}{2}(\Sigma-I) 
\end{align}

For MAP, the estimation can be considered as a regularized parameter optimization problem.

\subsubsection{Maximize the probability}
Our goal is to find the integral of
\begin{align}
\frac{\partial}{\partial m} \int_{x \in U_{\epsilon}(m)} f(x) d x & \!=\!\Sigma^{-1} \int_{z \in U_{\epsilon}(0)}(m\!+\!z\!-\!\mu) \!\cdot\! f(m\!+\!z) d z  \\
& =\Sigma^{-1} \int_{z \in U_{\epsilon}(0)}(m+z) \cdot f(m+z) d z\nonumber \\
& -\Sigma^{-1} \int_{z \in U_{\epsilon}(0)} \mu \cdot f(m+z) d z 
\end{align}

The first line is obtained by interchanging integration and differentiation. This integral is then solved separately.
\begin{equation}
\int_{z \in U_{\epsilon}(0)} f(m+z) d z=\int_{x \in U_{\epsilon}(m)} f(x) d x 
\end{equation}
\begin{equation}
\!=\!\frac{1}{2 \pi \sqrt{|\Sigma|}} \int_{x \in U_{\epsilon}(m)} \exp \left(\!-\!\frac{1}{2}(\mathbf{x}\!-\!\boldsymbol{\mu})^{T} \Sigma^{-1}(\mathbf{x}\!-\!\boldsymbol{\mu})\right) d x 
\end{equation}
\begin{equation}
=\frac{1}{2 \pi \sqrt{|\Sigma|}} \int_{x \in U_{\epsilon}(m)} \exp \left(-\frac{1}{2}(\mathbf{x}\!-\!\boldsymbol{\mu})^{T} \Lambda(\mathbf{x}\!-\!\boldsymbol{\mu})\right) d x 
\end{equation}
\begin{equation}
=\frac{1}{2 \pi \sqrt{|\Sigma|}} \int_{y \in U(m-\mu, \epsilon)} \exp \left(-\frac{1}{2} \mathbf{y}^{T} \Lambda \mathbf{y}\right)|T| d y 
\end{equation}
\begin{equation}
=\frac{1}{2 \pi \sqrt{|\Sigma|}} \int_{y \in U(m-\mu, \epsilon)} \exp \left(-\frac{1}{2} \mathbf{y}^{T} \Lambda \mathbf{y}\right) d y 
\end{equation}

The third line is obtained by the change of variables $T^{-1}(x-\mu)= y \Rightarrow d x=|T| d y$, and the following step is based on the fact that $T$ is an orthogonal matrix. To be more specific, the columns of $T$ are the eigenvectors of $\Sigma$, and the diagonal elements of $\Lambda$ are eigenvalues (i.e., $\Sigma=T \Lambda T^{T}$ ). Using a similar trick (or integration by parts), the following equation can be derived:
\begin{align}
& \int_{z \in U_{\epsilon}(0)}(m+z) \cdot f(m+z) d z=\int_{x \in U_{\epsilon}(m)} x f(x) d x \nonumber\\
& \!=\!\frac{1}{2 \pi \sqrt{|\Sigma|}} \int_{y \in U(m-\mu, \epsilon)}(T y\!+\!\mu) \exp \left(\!-\!\frac{1}{2} \mathbf{y}^{T} \Lambda \mathbf{y}\right) d y 
\end{align}

Bringing everything together gives the following result:
\begin{align}
& \frac{\partial}{\partial m} \int_{x \in U_{\epsilon}(m)} f(x) d x\nonumber \\
& \quad=\frac{\boldsymbol{\Sigma}^{-1} \boldsymbol{T}}{2 \pi \sqrt{|\boldsymbol{\Sigma}|}} \int_{y \in U(m-\mu, \epsilon)} y \exp \left(-\frac{1}{2} \mathbf{y}^{T} \Lambda \mathbf{y}\right) d y  \\
& \quad=\frac{T \Lambda}{2 \pi \sqrt{|\boldsymbol{\Lambda}|}} \int_{y \in U(m-\mu, \epsilon)} y \exp \left(-\frac{1}{2} \mathbf{y}^{T} \Lambda \mathbf{y}\right) d y 
\end{align}

The resulting expressions are intractable, as the integrals cannot be expressed in terms of elementary functions (one can verify that the double integral given in Eq. (40) involves a term of the error function, as calculating the cumulative distribution of a Gaussian distribution is required). Therefore, it may be more practical to use numerical or computational methods to approximate the derivatives rather than attempting to derive them analytically.

By observation, we find that the relation between $\frac{\partial f(x)}{\partial x}$ and $\frac{\partial f(m+z)}{\partial m}$ is subject to the constraint $x=m+z$.
\begin{align}
\frac{\partial f(\mathbf{x})}{\partial x} & =\frac{1}{2} \frac{\partial}{\partial \mu}\left[(\mathbf{x}-\boldsymbol{\mu})^{T} \Sigma^{-1}(\mathbf{x}-\boldsymbol{\mu})\right] \cdot f(\mathbf{x}) \\
& =\Sigma^{-1}(x-\mu) \cdot f(x) 
\end{align}
Similarly,
\begin{equation}
\frac{\partial f(\mathbf{m}+\mathbf{z})}{\partial m}=\Sigma^{-1}(m+z-\mu) \cdot f(m+z) 
\end{equation}
Thus, the following formula can be derived:
$$\frac{\partial}{\partial m} \int_{x \in U_{\epsilon}(m)} f(x) d x=\int_{x \in U_{\epsilon}(m)} \frac{\partial f(\mathbf{x})}{\partial x} d x$$

Next, we define the equation
\begin{equation}
\phi(m ; \epsilon)=\int_{x \in U_{\epsilon}(m)} f(x) d x
\end{equation}

\begin{figure}[!t]
\centering
\vspace{-0.1in}
\includegraphics[width=0.9\linewidth]{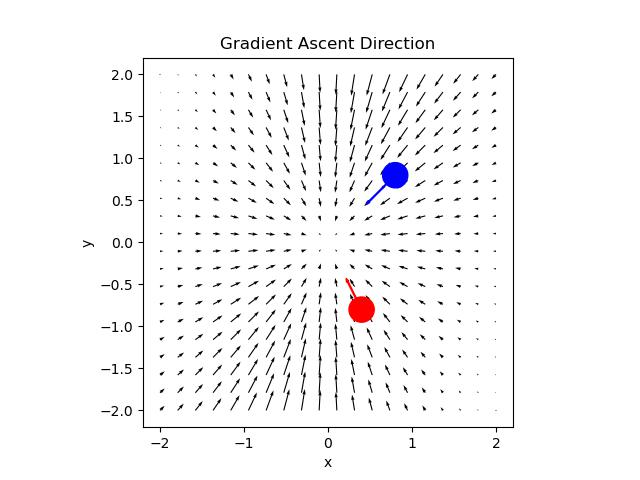}
\vspace{-0.1in}
\caption{Gradient ascent direction for the case of two search covers. The black short
arrows show the gradient descent directions. (For interpretation of the references to
color in this figure legend, the reader is referred to the web version of this article.)}
\label{f5}
\end{figure}

We define $\phi(m ; \epsilon)$, which indicates the chance of successfully making an insertion. To approximate the integral, the Monte Carlo sampling method could be applied. Furthermore, to obtain the maximal value of $\phi(m)$ numerically, we propose a numerical meta method.

Since only two dimensions are involved, the problem of the "Curse of Dimensionality" in the field of machine learning is not a concern. Furthermore, since the assumed distribution is simple enough, there is no need to implement methods like meta-dynamics to escape the socalled low-energy landscape. Thus, Monte Carlo seems to be a good choice.

On the other hand, since it is unnecessary to find a perfect solution, the best solution can be chosen empirically from several trials. The concept of the gradient descent method can be safely accepted. By combining the gradient descent method with the Monte Carlo method, it is possible to find a cover centered on $m$ that maximizes the value of $\phi(m)$. An example can be found in Fig. \ref{f5}.

Our final goal is to find $n$ covers that maximize the probability of successfully inserting the components. To avoid the collapse of the covers, a repelling force will be exerted between each pair of covers. The objective function is given as follows:
\begin{equation}
\Phi_{i}(m)=\phi\left(m_{i}\right)+\frac{\lambda}{n-1} \sum_{j \neq i} \max \left(\left\|m_{i}-m_{j}\right\|_{2}^{2}-\epsilon, 0\right)
\end{equation}
Alternatively, the repelling strength between each cover and the center of all covers can be added, yielding:
\begin{equation}
\Phi_{i}(m)=\phi\left(m_{i}\right)+\lambda \cdot\left\|m_{i}-\frac{1}{n} \sum_{j=1}^{n} m_{j}\right\|_{2}^{2} 
\end{equation}

In both Eqs. (45) and (46), the term $\lambda$ controls the strength of the penalty for distances between the centers of the covers. By solving the equations introduced, the function $F$ to be maximized is defined as
\begin{equation}
F=\sum_{i} \Phi_{i}(m) 
\end{equation}

Combining steps (1), (2), and (3), a complete meta search method is proposed, as shown in Algorithm 2. This method can be extended to any number of expected points for search. In the end, the centers of the covers will be found after the search, and their coordinates will be sent to the robot.

\begin{algorithm}[!t]
\caption{Meta Search Method}
\begin{algorithmic}[1]
\Require Assign the predefined insertion position as $m_0$, and initialize the prior distribution as $\mathcal{N}(m_0, I)$
\Repeat
    \State \textbf{Part 1: Maximize the Probability}
    \State Generate a sequence of candidates using Eq. (45) or (46)
    \Comment{$\triangleright$ Generate candidate points for subsequent steps}
    \State Sort the candidates by the probabilities of being inserted
    \Comment{$\triangleright$ Increase the likelihood of successful insertion}
    \While{not inserted}
        \State Try the next candidate point
        \Comment{$\triangleright$ If insertion is unsuccessful, move to the next candidate}
    \EndWhile
    \State \textbf{Part 2: Distribution Update}
    \State Update the posterior distribution based on the new point $x_t$, using Equation Eq. (26) or (28).
\Until{All components are successfully inserted}
\end{algorithmic}
\end{algorithm}

\begin{figure}[!t]
  \centering
  \includegraphics[width=1.0\linewidth]{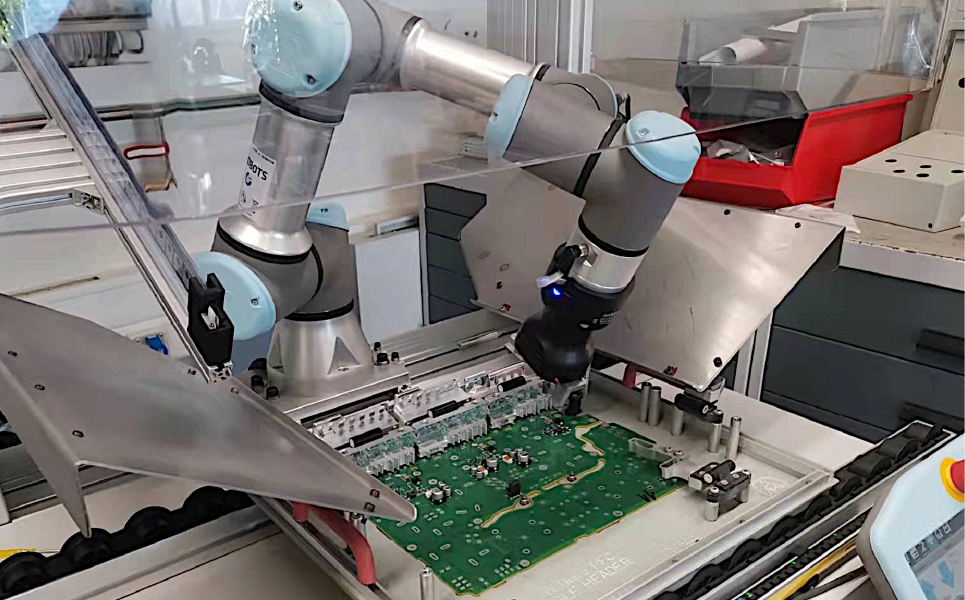}
    \vspace{-0.8cm}
  \caption{The assembly unit with one typical odd-form component.}
  \label{f6}
\end{figure}

%% file: system.tex
\section{Experiment}
\subsection{Experiment configuration}
The search method is implemented on a Universal Robot 3e (UR3e). The UR robot is integrated into a flexible assembly unit \cite{36} (see Fig. \ref{f6}). The assembly unit is equipped with a safety light curtain miniTwin4 for safe interaction with operators during assembly \cite{37}. If a human enters the robot's working area, the light curtain is triggered, which subsequently halts the robot's operation. If the IGBT cannot be successfully inserted into its corresponding holes after reaching the maximum allotted search cycles, human intervention is required to halt the assembly process. A bare PCB is placed on a carrier for each assembly cycle. The carrier, along with the PCB, is transported to the assembly unit via conveyors. In each cycle, the robot picks up one odd-form component and moves it to the initialized insertion position. At this point, one of the search methods is selected, and the insertion process is executed.

\begin{figure}[!t]
\centering
\includegraphics[width=1\linewidth]{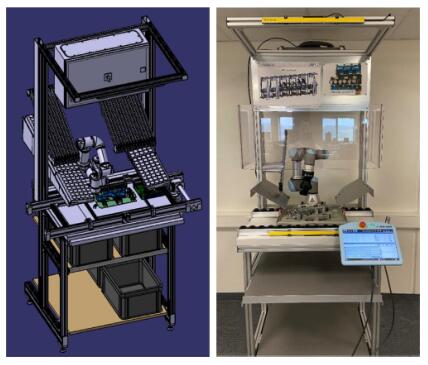}
\vspace{-0.1in}
\caption{Assembly model and real experimental environment.}
\vspace{-0.1in}
\label{f7}
\end{figure}

In addition to the robotic system, the laboratory also had a highperformance computer configured to run the meta-search algorithm. The computer specifications included an Intel Core i7 processor with 16 GB of RAM and a NVIDIA GeForce RTX 3060 graphics card. This configuration provided sufficient computational power to process the complex algorithms and data required for the meta-search method. The assembly model and the real experimental environment are shown in Fig. \ref{f7}.

\subsection{Experiment design}
There are two types of odd-form components according to their insertion characteristics: polar and non-polar odd-form components. The polar odd-form component can be inserted when its insertion direction and position are accurately set, whereas the non-polar oddform components only require the correct insertion position. Therefore, the search strategy is suitable for both polar and non-polar odd-form components. A polarity detection submodule should be additionally incorporated to insert polar odd-form components.

\begin{figure}[!t]
\centering
\includegraphics[width=1\linewidth]{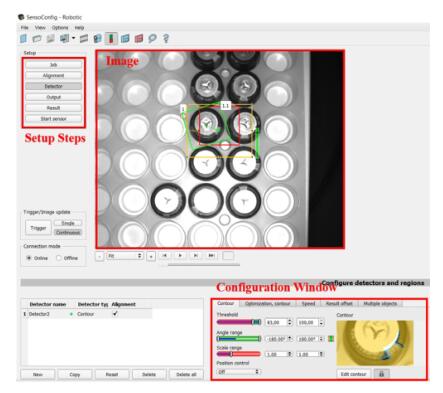}
\caption{Comparison of linear strategy and initial point optimization strategy.}
\label{f8}
\end{figure}

\begin{figure}[!t]
\centering
\includegraphics[width=1\linewidth]{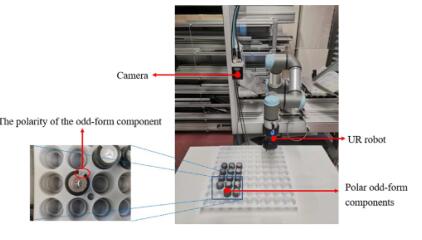}
\caption{The assembly unit with the vision submodule.}
\label{f9}
\end{figure}

The polarity direction can be determined by configuring the integrated software, which simplifies the setup process and increases detection accuracy. The configuration interface of the software is shown in Fig. \ref{f8}. The detected polarity direction is returned as the gripping pose for the UR robot. The UR robot grips each polar odd-form component from different gripping poses and inserts it in a specific direction. The robot relies on the aforementioned search strategies to accurately and quickly insert the component. The prototype of the assembly unit with the vision submodule is displayed in Fig. \ref{f9}.

During the experiment, the assembly time is used as the metric for evaluating the efficiency of the search methods. All search methods employed in the experiments include:

\noindent{\bf (1) Line Search:} The robot implements the linear search method solely and performs a search for the insertion hole along a predefined direction to ensure the holes can be consistently located after multiple attempts.

\noindent{\bf (2) Spiral Search:} The robot implements the spiral search method with initial point optimization. In this case, the robot performs a search along a spiral path.

\noindent{\bf (3) Hybrid Search:} The robot alternates between performing a one-step line search and then a one-step spiral search.

\noindent{\bf (4) Meta Search:} The robot assembles the PCB using the proposed meta search method.

For each experiment, the robot performs continuous assembly of $\mathbf{10}$ different PCBs of the same type. In each individual PCB, a total of $\mathbf 9$ IGBTs need to be inserted. The assembly process is concluded when the insertion hole is successfully located or the maximum allotted search time is reached. It is worth noting that the 7th hole position in each PCB is abnormal due to errors in the manufacturing process of the PCBs.

The line search method, spiral search method, and hybrid search method require optimizing the initial point before each cycle. More precisely, these methods leverage an average approach before each cycle, meaning that for these three search methods, the starting point for each assembly cycle is computed as the mean of all preceding positions. In contrast, the meta search method operates under the assumption that the insertion holes follow a Gaussian distribution, which is continually updated. As a result, there is no need for additional computations to determine the initial position, presenting an inherent advantage of the meta search method.

\subsection{Overall performance}
\begin{figure}[!t]
\centering
\includegraphics[width=1\linewidth]{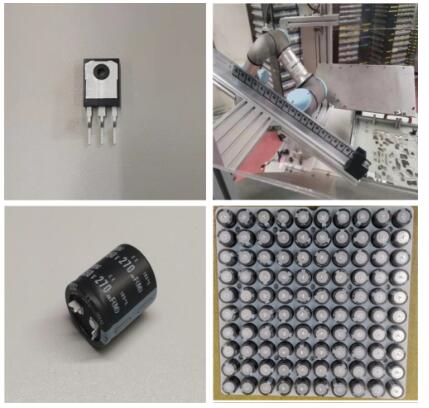}
\caption{IGBT and Capacitor Components in assembly.}
\vspace{-0.1in}
\label{f10}
\end{figure}

We evaluate the performance of the proposed methodology in a realistic robotic assembly setup and compare it with other search methods. We conducted assembly experiments on two types of electronic components, namely IGBT and Capacitor, independently. The images of these two components, IGBT and Capacitor, along with their preassembly states, are illustrated in Fig. \ref{f10}. The IGBTs arrive from the supplier packaged in plastic tubes. This packaging design simplifies the development of a material feeding system. On the other hand, the capacitors have a cylindrical shape with leads located on the bottom and polarity marked on the top. The experimental results for each component are presented in Table \ref{t1} and \ref{t2}, respectively. The two Tables present a comprehensive comparison of the assembly cycle times achieved by four different search methods: Line Search, Spiral Search, Hybrid Search, and our proposed Meta Search. It is evident from the data that the Meta Search method consistently outperforms the others across all ten assembly cycles, achieving the lowest average assembly time of 42.6 s. This represents a substantial improvement over the Linear Search (106.6 s average), which struggled to adapt and optimize its search path. The Spiral Search (66.0 s average) and Hybrid Search (62.3 s average) showed moderate improvement over Linear Search but still lagged behind the Meta Search method.

\begin{table*}[h] 
\renewcommand\arraystretch{1.3}
\centering
\caption{Comparison of positioning ability for assemble task (IGBT). }
\label{t1}
 \vspace{-0.2cm}
\resizebox{\linewidth}{!}{
    \begin{tabular}{c|c|c|c|c|c|c|c|c|c|c|c}
    \thickhline
    \multirow{2}{*}{Method} & \multicolumn{10}{c|}{Assemble Cycle Time(s)}    & \multirow{2}{*}{\makecell[c]{Average\\Assemble Time}} \\ \cline{2-11}
     & C1  & C2  & C3  & C4  & C5  & C6  & C7  & C8  & C9  & C10 &  \\ \hline
    Line Search  & 106.2 & 105.2 & 106.0 & 106.9 & 107.7 & 104.8 & 109.2 & 108.5 & 105.6 & 109.1 & 106.6 \\\hline
    Spiral Search  & 102.7 & 72.1  & 64.6  & 58.3  & 62.8  & 60.4  & 63.8  & 59.9  & 57.3  & 61.1  & 66.0  \\\hline
    Hybrid Search  & 78.1  & 64.8  & 62.6  & 63.6  & 60.8  & 61.5  & 59.2  & 57.7  & 56.8  & 58.5  & 62.3  \\\hline
    \textbf{Meta Search }  & 73.5  & 48.8  & 40.9  & 40.3 & 42.0  & 38.7  & 36.4  & 36.6  & 37.7  & 35.6  & 42.6  \\ \thickhline
    \end{tabular}
}
\vspace{-0.4cm}
\end{table*}

\begin{table*}[h] 
\renewcommand\arraystretch{1.3}
\centering
\caption{Comparison of positioning ability for another assemble task (Capacitor).}
\label{t2}
 \vspace{-0.2cm}
\resizebox{\linewidth}{!}{
    \begin{tabular}{c|c|c|c|c|c|c|c|c|c|c|c}
    \thickhline
    \multirow{2}{*}{Method} & \multicolumn{10}{c|}{Assemble Cycle Time(s)}    & \multirow{2}{*}{\makecell[c]{Average\\Assemble Time}} \\ \cline{2-11}
     & C1  & C2  & C3  & C4  & C5  & C6  & C7  & C8  & C9  & C10 &  \\ \hline
 Line Search & 120.5 & 118.7 & 119.3 & 121.1 & 122.4 & 117.9 & 120.6 & 120.8 & 119.2 & 124.0 & 120.3 \\
\hline Spiral Search & 98.3 & 85.6 & 82.1 & 79.4 & 81.2 & 78.9 & 79.5 & 77.3 & 76.1 & 79.7 & 81.6 \\
\hline Hybrid Search & 92.4 & 80.7 & 77.6 & 78.2 & 76.5 & 75.8 & 74.9 & 73.1 & 72.3 & 74.4 & 76.3 \\
\hline {\bf Meta Search} & 65.2 & 54.3 & 49.8 & 48.7 & 47.9 & 46.2 & 45.1 & 44.3 & 43.9 & 43.5 & 47.1 \\

\thickhline
    \end{tabular}
}
\vspace{-0.4cm}
\end{table*}

Linear Search method relies solely on a predefined linear path, making it inflexible and prone to errors when the actual assembly position deviates from this path. As a result, the assembly time remains relatively constant throughout the experiment, hovering around an average of 106.6 s for the first set of trials and 120.3 s for the second set, indicating a significant lack of adaptability. The method's inability to adjust to deviations results in consistently high assembly times, regardless of the specific assembly cycle or task.

Spiral Search and Hybrid Search demonstrate an improvement over Linear Search by progressively optimizing the initial search point. The spiral path of Spiral Search and the alternating strategy of Hybrid Search allow for a more exhaustive search space exploration, resulting in average assembly times of 66.0 s and 62.3 s respectively for the first task, and 81.6 s and 76.3 s for the second task. However, both methods still rely heavily on empirical averages or predefined patterns to guide their search. This reliance limits their adaptability to unforeseen deviations, such as those caused by manufacturing errors or variations in part dimensions.

The key advantage of our proposed Meta Search method lies in its ability to perceive assembly positions as a Gaussian distribution that continuously updates based on past observations. This learning capability enables the Meta Search method to dynamically adjust its search strategy in real-time, resulting in a significant reduction in assembly time. For the first task, the Meta Search method achieved an average assembly time of 42.6 s, and for the second task, it achieved an average of 47.1 s. As the experiment progresses, the method becomes increasingly efficient at predicting and locating the optimal insertion holes, thanks to its continuous learning and adaptation.

From an algorithmic perspective, the Meta Search method's advantage stems from its ability to leverage past data to inform future decisions. The method uses a probabilistic model to represent the distribution of assembly positions and updates this model as new data is collected. As more data is accumulated, the model becomes more accurate, leading to faster and more reliable assembly performance. This continuous learning and adaptation are crucial in real-world applications, where manufacturing errors, deviations from ideal conditions, and variations in part dimensions are inevitable.

In contrast, the other methods - Linear Search, Spiral Search, and Hybrid Search - either rely on fixed patterns (Linear Search) or empirical averages (Spiral and Hybrid Search), limiting their ability to adapt and improve over time. The Meta Search method, with its probabilistic model and continuous learning capability, offers a significant advantage in terms of adaptability, accuracy, and efficiency in robotic assembly tasks.

\subsection{Study of self-adaption ability}
\begin{figure}[!ht]
  \centering
  \includegraphics[width=1.0\linewidth]{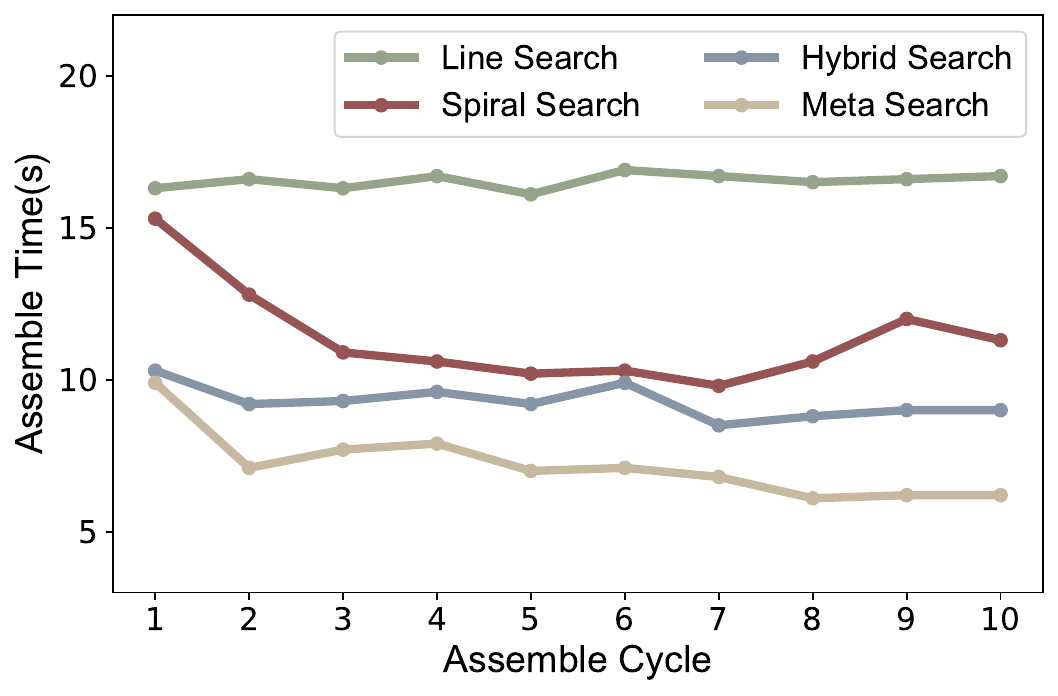}
    \vspace{-0.8cm}
  \caption{Assemble time for abnormal position.}
      \vspace{-0.2cm}
  \label{f11}
\end{figure}

In exploring deeper into the self-adaption prowess of our search methodologies, we focus on the 7th hole position across PCBs, identified as an anomaly due to its more scattered distribution. This section meticulously examines how each approach navigates this irregularity. The experimental results are shown in Fig. \ref{f11}.

The linear search method, by its very nature, follows a predetermined, straight-line path to locate the holes. As a result, when confronted with the irregular distribution of the 7th hole, it struggles to adapt, maintaining a consistently high assembly time throughout the experiments. This lack of flexibility underscores the method's limitation in handling dynamic or unpredictable assembly scenarios.

On the other hand, the spiral search and hybrid search methods demonstrate a notable improvement in assembly speed over time, particularly for the 7th hole. This is attributed to their underlying mechanisms that leverage an average or median position of previously encountered holes to narrow down the search range. As the experiments progress, these methods gradually refine their search strategy, leading to faster assembly speeds. However, their reliance on statistical averages imposes inherent constraints, limiting their adaptability to extreme or highly irregular cases.

The meta search method stands out in this test of adaptability. Its superior performance stems from its intricate algorithm that treats hole positions as dynamic Gaussian distributions, continuously updating these distributions based on real-time observations. When confronted with the abnormal 7th hole, the meta search method swiftly adjusts its search strategy, incorporating the new data into its statistical models. This allows it to dynamically refine its predictions and adapt to the irregular distribution, achieving the fastest assembly time among all methods. Furthermore, this adaptability extends beyond mere irregular hole positions; it suggests that the meta search method is well-suited for tackling even more complex scenarios, such as holes with non-standard shapes (e.g., ellipses), demonstrating its versatility and robustness in robotic assembly tasks.

\subsection{Study of regularization factor}
In this comprehensive study of the regularization factor $\lambda$ within the meta search method's framework, we explore into its pivotal role in balancing the trade-off between exploring diverse candidate assembly positions and punishing those that are less favorable. Eqs. (45) and (46) embody this mechanism, where $\lambda$ acts as a fine-tuning knob, adjusting the severity of the penalty imposed on suboptimal positions.

\begin{figure*}[!t]
\includegraphics[width=0.50\textwidth]{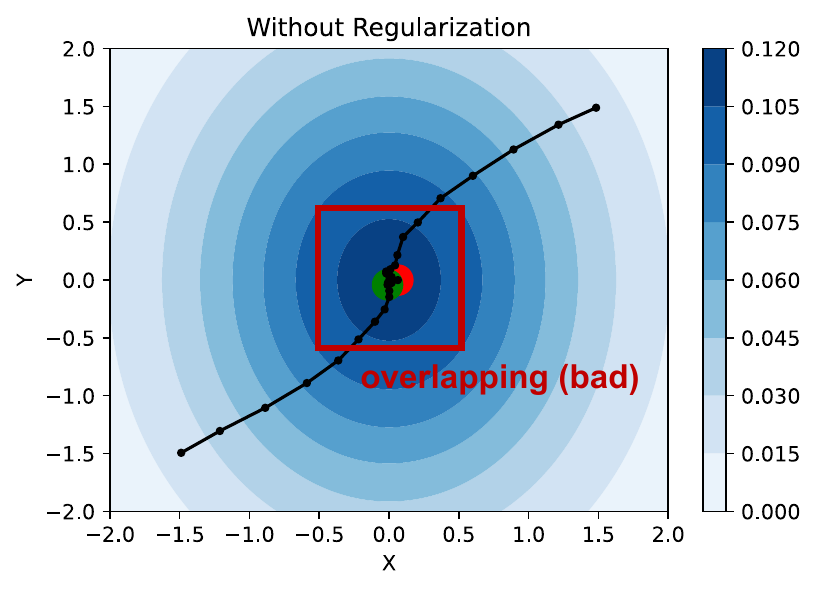}
\includegraphics[width=0.50\textwidth]{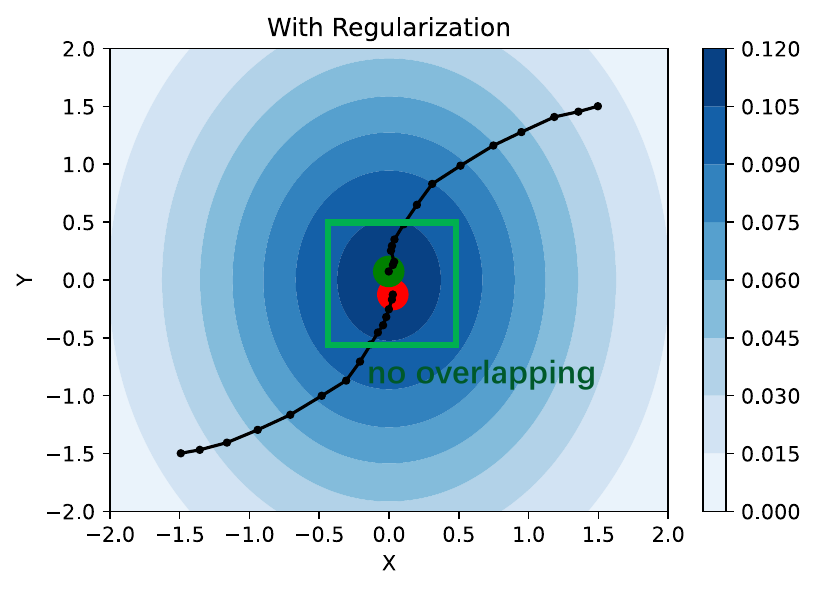}
\vspace{-0.7cm}
\caption{Search the cover(s) of maximum probability (\textbf{Left:} search without regularization and \textbf{Right:} search with regularization). Notice that without any regularization the covers will tend to collapse.}
\label{f12}
\end{figure*}

Fig. \ref{f12} vividly illustrates the stark contrast between the meta search method's performance with and without regularization. In the absence of regularization, the figure reveals a chaotic landscape of candidate assembly positions, where significant overlaps occur, hindering the algorithm's ability to efficiently converge on the optimal solution. This leads to a notable slowdown in the assembly process, underscoring the critical need for a well-calibrated regularization factor.

To quantify the impact of $\lambda$, we conducted a meticulous ablation study, tabulating the average assembly time across ten distinct PCBs as a function of various $\lambda$ values. The experiment result is listed in Table \ref{t3}. The results align with our theoretical expectations. When $\lambda$ is set too high, the algorithm becomes overly conservative, hesitating to explore promising but slightly deviated positions, leading to divergence and suboptimal performance. Conversely, a $\lambda$ that is too low fails to adequately discourage suboptimal candidates, resulting in a collapse of the search space and a subsequent degradation in assembly speed.

\begin{table}[!h]
\centering
\caption{Ablation study of regularization factor $\lambda$}
\vspace{-0.2cm}
\label{t3}
\renewcommand\arraystretch{1.3}
\begin{tabular}{c|c|c|c|c|c}
\thickhline
Regulization Fator $\lambda$ & $1e^{-2}$ & $1e^{-3}$& $1e^{-4}$ & $1e^{-5} $&  $ 1e^{-6} $ \\
\hline
Average Assemble Time(s)  & 89.4  &85.2  & 42.5       & 56.9     & 72.5      \\   
\thickhline
\end{tabular}
\end{table}

Our findings highlight a sweet spot at $\lambda=1 e^{-4}$, where the regularization factor strikes the perfect equilibrium between exploration and exploitation. This optimal setting enables the meta search method to swiftly navigate through the complex landscape of candidate assembly positions, efficiently identifying and converging on the most favorable options, thereby minimizing the overall assembly time.

%% file: experiment.tex
\section{Conclusions and future work}
In this study, we conducted a comprehensive analysis of the metasearch method for optimizing the assembly process of printed circuit boards. We proposed a novel meta-search method for precise robotic positioning, which has self-learning capabilities and can adapt to new assembly tasks in constrained environments that are unsuitable for integrating vision systems. We also examined various aspects of the methodology, including its sensitivity to parameters, performance metrics, and potential for improvement. The effectiveness and self-adaptive ability of this method were demonstrated through experiments in a real robotic assembly environment. The main achievements of this paper are summarized in the following aspects:

\begin{itemize}
  \item {\bf Vision-Independence Assembly:} The development of a vision-free and model-agnostic meta-method for robotic position error compensation represents a significant step forward. This approach's independence from vision systems and predefined models unlocks new possibilities for industrial applications. It streamlines integration into existing production lines, reduces hardware dependencies, and fosters cost savings. For researchers and practitioners alike, it encourages the exploration of alternative feedback-driven control strategies that can generalize across a wide range of robotic tasks.
  \item {\bf Meta Learning Method:} A Catalyst for Continuous Improvement: By empowering robots with the ability to learn and adapt to various position errors, this method introduces a new level of intelligence to robotic systems. Its self-learning and self-adaptive nature allow robots to continually refine their positioning accuracy as they encounter new scenarios, minimizing downtime due to errors and boosting overall productivity. Moreover, it paves the way for robots to autonomously handle increasingly complex tasks, reducing human intervention and improving workplace safety.
  \item {\bf Practical Validation:} The successful implementation of this method in a robotic assembly line for intricate odd-form electronic components serves as a testament to its practicality and readiness for industrial deployment. This milestone underscores the effectiveness of the approach in addressing real-world challenges. By improving accuracy, reducing waste, and enhancing overall productivity, it contributes to more sustainable and cost-effective production methods. Furthermore, this practical validation encourages wider adoption of the technology, fostering collaboration between industry and academia to further refine and expand its capabilities.
\end{itemize}

Three interesting future research directions also exist: {\bf Robotic Positioning for Deformable Components.} Robotic positioning for deformable components presents a frequently encountered yet challenging scenario, primarily due to the susceptibility of these components to damage. Repeated assembly attempts often lead to deformation, which in turn hinders the assembly process. Enhancing the proposed meta-search method to incorporate adaptive strategies that minimize deformation and maximize assembly success rates would be highly beneficial. This could involve refining the search algorithms to better handle compliant materials and incorporating feedback mechanisms that adjust the robotic movements based on real-time deformation assessments.

{\bf Dynamic Robotic Positioning for Movable Components.} Another promising area of research lies in extending the meta-algorithm to support dynamic positioning for movable components. By enabling the method to accurately adapt to the target object's motion trajectory and simultaneously compensate for positioning errors, significant contributions to flexible manufacturing can be made. This would involve integrating advanced motion tracking and prediction techniques into the meta-algorithm, allowing it to anticipate and react to changes in the component's position in real-time.

{\bf Collaborative Human-Robot Assembly.} Finally, exploring collaborative human-robot assembly techniques could also be a fruitful area of research. By combining the precision and efficiency of robots with the judgment and adaptability of human operators, it may be possible to create a more robust and flexible assembly process. For example, human operators could manually adjust components that the robot has difficulty positioning, while the robot could handle the more repetitive and precise tasks. Such a collaborative approach could help to minimize collision issues and improve overall assembly quality.

%% file: conclusion.tex
\vspace{-0.1in}
\section*{\textbf{CRediT authorship contribution statement}}
\emph{The credit was deliberately mis-attributed in the accepted version and now corrected.} Jieyang Peng: Writing - original draft. Dongkun Wang: Experiment. Junkai Zhao: Software. Yunfei Teng: Methodology. Andreas Kimmig: Guest authorship. Xiaoming Tao: Partial funding acquisition and guest authorship. Jivka Ovtcharova: Guest authorship.

Y. Teng conducted independent research and authored Section 3 Methodology, including all related algorithms, descriptions, and the creation of all associated figures. Unauthorized use of this material has been reported in the journal article \textit{(DOI:10.1016/j.jmsy.2024.11.009)} and the patent \textit{(CN 118960772 A)}. His contributions to the manuscript were strictly limited to meta-method design, theoretical analysis, and the writing and revision of the overall paper, with no involvement beyond these stated components.

The other parts of the work were partially supported by EU H2020 Research and Innovation Program under the Marie Sklodowska-Curie Grant Agreement (Project-DEEP, Grant No. 101109045), National Key R\&D Program of China (No. 2017YFE0101400). National Key R\&D Program of China with Grant number 2018YFB1800804, the National Natural Science Foundation of China (Nos. NSFC 61925105, and 62171257) and Tsinghua University-China Mobile Communications Group Co., Ltd. Joint Institute and the Fundamental Research Funds for the Central Universities, China (No. FRF-NP-20-03). The research is also supported by the German KIT- internal research project Wertstromkinematik (Value Stream Kinematics) and the German Federal Ministry of Education and Research (BMBF) (project AITT, AI-assisted Technology Transfer, No. 03LB3058B).

\vspace{-0.04in}
\section*{\textbf{Declaration of competing interest}}
\vspace{-0.03in}
This manuscript is incomplete at the time of acceptance by the \emph{Journal of Manufacturing Systems}, including unchecked algorithmic components and unverified experimental results. We notified the journal editor \emph{(\url{ihuiw@kth.se})} and publisher \emph{(t.bazuin@elsevier.com)} of these issues; however, no correction or amendment has been made to the accepted version. Thanks to them, the reported experiments are not reproducible and the manuscript’s acceptance appears to have relied primarily on limited presentation materials (e.g., selected figures) rather than a complete and verifiable technical contribution.